# Compact Convolutional Neural Network Cascade for Face Detection

Kalinovskii I.A.
Tomsk Polytechnic University
Tomsk, Russia
kua_21@mail.ru

Spitsyn V.G.
Tomsk Polytechnic University
Tomsk, Russia
spvg@tpu.ru

## Abstract

*The problem of faces detection in images or video streams is a classical problem of computer vision. The multiple solutions of this problem have been proposed, but the question of their optimality is still open. Many algorithms achieve a high quality face detection, but at the cost of high computational complexity. This restricts their application in the real-time systems. This paper presents a new solution of the frontal face detection problem based on compact convolutional neural networks cascade. The test results on FDDB dataset show that it is competitive with state-of-the-art algorithms. This proposed detector is implemented using three technologies: SSE/AVX/AVX2 instruction sets for Intel CPUs, Nvidia CUDA, OpenCL. The detection speed of our approach considerably exceeds all the existing CPU-based and GPU-based algorithms. Because of high computational efficiency, our detector can processing 4K Ultra HD video stream in real time (up to 27 fps) on mobile platforms (Intel Ivy Bridge CPUs and Nvidia Kepler GPUs) in searching objects with the dimension 60×60 pixels or higher. At the same time its performance weakly dependent on the background and number of objects in scene. This is achieved by the asynchronous computation of stages in the cascade.*

## 1. Introduction

The need to identify people on millions of photos uploaded daily to social services has led to significant progress in solution of the problem of detecting faces. New methods have invariance with respect to pose and face expression. Moreover, they are also capable of operating in a complex illumination and strong occlusion. However, many algorithms that demonstrate outstanding performance on face detection benchmark have very high computational complexity. This circumstance prevents their use for video analysis.

The object of our interest is megapixel video analytics systems that require fast and accurate face detection algorithms. Such systems run on the equipment, which computing power is often greatly limited because of the increasing demands for a compact form factor and cost. Because of it, the increase in frame rate or frame resolution is often carried out at the expense of the performance of detection (large size objects search only, frames skipping, etc.). Moreover, the use of cameras capable of shooting video with a 4K Ultra HD resolution increases the amount of generated data several times. In conditions, when it is impossible to reduce the search area (for example, by motion analysis), even optimized detectors based on Viola-Jones method are not able to operate effectively with such video stream resolution.

Despite the fact, that development of modern face detection methods is moving towards increase of invariance with respect to the head position and the occlusion, we consider only a particular problem of frontal faces detection. Our goal is to achieve high performance of detection at low computational complexity of the detector, which is difficult to achieve when dealing with this problem in the most general problem definition. At the same time, the frontal position of a person in relation to the camera is natural for many video analytics system use case. That is why these detectors are so popular in practical applications.

In this paper, we present a frontal face detector based on the cascade of convolutional neural network (CNN) [12] with a very small number of parameters. Due to the natural parallelism, a small number of cascade stages and low-level optimization, it is capable of processing real-time 4K Ultra HD video stream on mobile GPU when searching for faces with the size of 60×60 pixels or higher, and the same time it is 9 times faster than the detector based on Viola-Jones algorithm in the OpenCV implementation (http://opencv.org). Despite the compact CNN architecture, test results on Face Detection Data Set and Benchmark (FDDB) dataset [8] show that our CNN cascade is comparable in performance with the state-of-the-art frontal face detectors. It surpasses any existing CPU and GPU algorithms in speed.

## 2. Related Work

It is not possible to build a simple and rapid detector with high precision and response to all the possible face images variations because of the big interclass variance, the variety of ambient light conditions, as well as the complex structure of the background. The standard approach to solving this problem is to use different models for each pose of the head [19, 32, 35]. It has been shown recently that, due to the strong generalization capability, deep CNNs can study the whole variety of two-dimensional projections of a face within the limits of a single model [3, 16]. However, the fact, that the proposed CNN architectures contain several million parameters, makes them unsuitable for use in low-power computing devices. Methods, based on deformable part models [19, 33], template comparison based models [15] and 3D face models [1], are also not able to work with HD video streams in real time, even to solve the problem of frontal faces search only. Detectors, which use manually designed features to describe objects and cascades of boosting classifiers to detect them, remain the best solution in terms of processing speed [26].

Many different descriptors were proposed to describe facial features. The most famous ones are rectangular Haar-like features [30], which have shown to be effective for building frontal face detectors and to have high extraction rate using the integral image. Textural MCT [4, 27] and LBP [29] features, which code pixel intensity in the local domains, have invariance with respect to monotonic light change. LBP, in combination with HOG features [24], demonstrated good generalizing properties and they are able to process complex non-facial images better in comparison with the Haar-like features. B. Jun et al. [10] proposed LGP and BHOG features, built on the principles of LBP. LGPs are resistant to local changes of light along the borders of the objects and BHOGs are resistant to local pose changes.

Multidimensional SURF descriptors [17] in combination with the logistic regression make it possible to prepare cascades containing only a few hundreds of weak classifiers. Because of it, SURF cascades exceed the speed of Haar cascades, which typically consist of thousands of weak classifiers. Simple comparison of pixel intensities can also be used for faces detection [2, 18]. Proposed in [18] detector has high execution speed, since it does not require any additional processing, including the construction of the image pyramid.

Usually, boosting cascades are trained using grayscale images for the solution of face detection problem. M. Mathias et al. [19] and B. Yang et al. [35] used combinations of different channels (grayscale, RGB, HSV, HOG descriptors and other) for classifiers training. Taking into account both color and geometric information improved the performance of face detection on a complex background.

Recently, H. Li et al. [16] have built a CNN cascade, which has the highest speed among the multi-view face detectors. Just as in Viola-Jones algorithm, a simple CNN was used for coarse image scanning, while more complex models carefully estimated each selected region. However, despite the significant reduction of the computational complexity (in comparison with the single CNN model [3]), this detector is still not capable of processing HD video stream even on powerful GPU.

It will be shown further that the CNN cascade may surpass boosting cascades not only in performance but also in speed for solution of the frontal faces detection problem. The CNN densely extracts high-level features directly from the raw data, without requiring any preliminary processing, apart from building the image pyramid. Besides, the CNN calculation algorithm can be easily vectorized using SIMD instructions of CPU and it can be perfectly adapted for massively parallel architecture of GPU.

## 3. Compact CNN cascade

Similarly to [16], we use the CNN to build a cascade detector of frontal faces. This work is based on the following key ideas:

1) *A small number of cascade stages*. The CNN cascade has only 3 stages. For example, the shortest boosting cascade consists of 4 stages and uses the MCT descriptors for extraction of facial features [4].

2) *Compact design of the CNN architectures*. The total number of feature maps in all CNNs is 355 (in [16] it amounts up to 1,949), but a samples with a smaller variation of face images was applied to train the model.

3) *Asynchronous execution of the stages*. A particular detector design makes it possible to execute the second and the third stages of the cascade in parallel with the first one on different processing units. Due to the fact that 99.99% of sliding window positions rejected already on the first operation stage, in this mode the detector is capable of processing video frames in constant time, regardless of the content of the image.

4) *Optimization*. During the detector implementation, three technologies were used: SIMD expansion of CPU, CUDA and OpenCL. The SIMD (CPU) and CUDA implementations of the CNN were optimized for each used network architecture. Giving up on the traditional approach of the CNN calculation through the organization of the stack of layers, combined with the assembler level optimization, made it possible to achieve the performance of code execution close to the peak hardware performance.

### 3.1. CNN structures

The CNN architectures composing a cascade are shown in Figure 1. Each CNN solves the problem of a background/face binary classification and contains 797, 1,819 and 2,923 parameters respectively. Similarly to the Convolutional Face Finder [5] architecture, the lack of fully-connected layer gives a 50% increase in speed of a forward propagation procedure. Convolution stride is 1 pixel, pooling stride is 2 pixels. Rational approximation of a hyperbolic tangent is used as an activation function:

$$f(x) = 1.7159 \cdot \tanh\left(\frac{2}{3}x\right),$$

$$\tanh(y) \approx \operatorname{sgn}(y)\left(1 - \frac{1}{1 + |y| + y^2 + 1.41645 \cdot y^4}\right). \quad (1)$$

The relative error of the following approximation does not exceed 1.8% on the entire number axis and only 11 instructions are required to calculate it. Popular ReLU functions turned out to be less efficient in our experiments. It should be noted that the $CNN_1$ contains the smallest number of filters in comparison with previously proposed network architectures for face detection [3, 5, 16, 23].

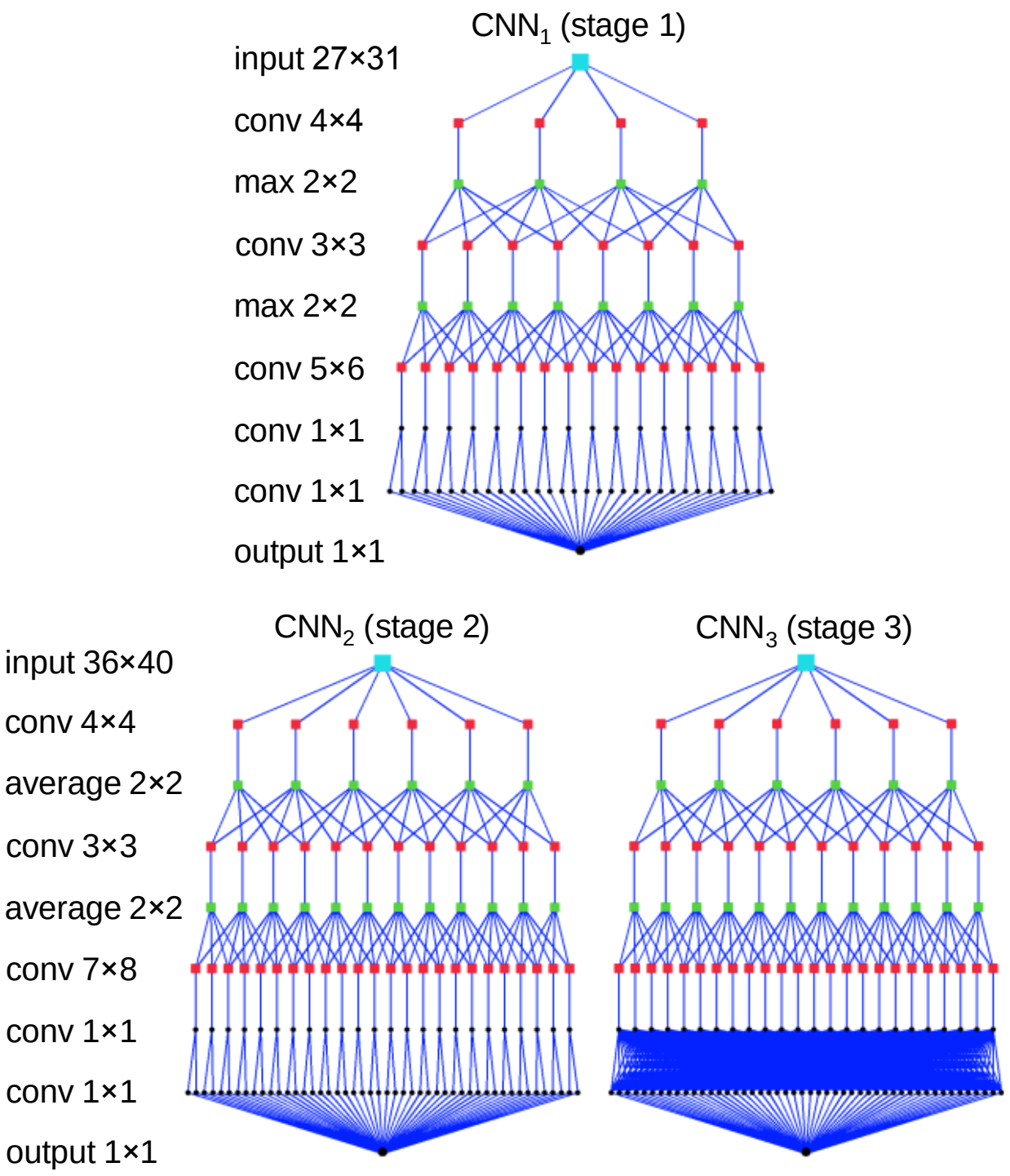

Figure 1: CNN structures.

### 3.2. Training process

For the CNN training aligned face images were taken from YouTube Faces Database [31]. This data set contains face tracks of 1,595 people cut-out from 3,425 videos (Fig. 2). Background images were selected from random YouTube videos in several stages during the preparation of models. Face areas (eyes, nose, etc.) were also added to the negatives. The total volume of training set consisted of slightly more than one million grayscale images (433 thousand of positive examples and 585 thousand of negative ones).

The experiments were carried out with simple models which have a small number of parameters. We tried to find the minimal configuration of the CNN, that is able to classify the test set with an error of no more than 0.5%. For the CNN training a Levenberg-Marquardt algorithm was used [36].

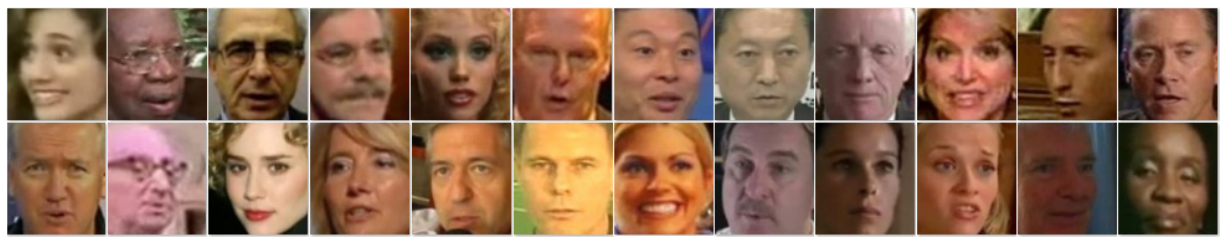

Figure 2: Example images from the training set.

### 3.3. Detector design

The detector design is shown in Figure 3. The first CNN densely scans in series each image of the pyramid. The responses in the output network layer correspond to the positions of the scanning window with a size of 27×31 pixels during its uniform motion with a 4 pixel step. The coordinates of the windows, where the CNN response exceeded the predetermined threshold $T_1$, are transmitted to the selective unit for further analysis of these regions of the image. Even when $T_1 = 0$, at this stage more than 99.99% from the total number of positions of the window at all pyramid levels are rejected. For comparison, the first-stage of Haar cascade of Viola-Jones [30] is able to reject only 50% negative samples, MCT cascade [4] – 99%, SURF cascade [17] – 95%, CNN cascade [16] – 92%.

Pre-processing and classification of image region are carried out in the selective unit, and then the final decision about their belonging to the faces class is made. At the step of pre-processing, the analyzed region is read from the original grayscale image together with certain neighborhood and scaled to the size of 51×55 pixels. Then, the equalization of its histogram and mirror reflection with respect to the vertical axis are carried out. Illumination alignment enhances the response of the CNN on the shaded faces and effectively suppresses false detections. The use of mirror reflection also reduces the response to the complex non-facial images.

On the second step, a region classification is carried out with the second and third stages of the cascade. The output of each CNN is a response map with a 5×5 size. Additional classification in the region neighborhood is necessary to prevent the loss of response due to incorrect positioning of the face in the scanning window. The decision about the type of the region is made on the basis of the number of responses $K_{nn}$ of each classifier that exceeds the predetermined threshold $T_2$:

$$\delta = \left(K_{nn}^{CNN2} \geq T_{nn} \wedge K_{nn}^{CNN3} > 0\right) \vee \left(K_{nn}^{CNN2} > 0 \wedge K_{nn}^{CNN3} \geq T_{nn}\right),$$

$$\text{decision rule} = \begin{cases} \delta = 1, \; face \\ \delta = 0, \; no \;\; face \end{cases}. \quad (2)$$

Discrete parameter $T_{nn}$ makes it possible to control the number of detector false alarms more robustly, as compared to the thresholds $T_1$ and $T_2$. If the response of the $CNN_2$ does not agree with the decision rule, then the further analysis of the region stops.

The last stage of the pipeline detector is Non-Maximum Suppression (NMS) algorithm, which aggregates the found regions to form the resulting areas of faces localization.

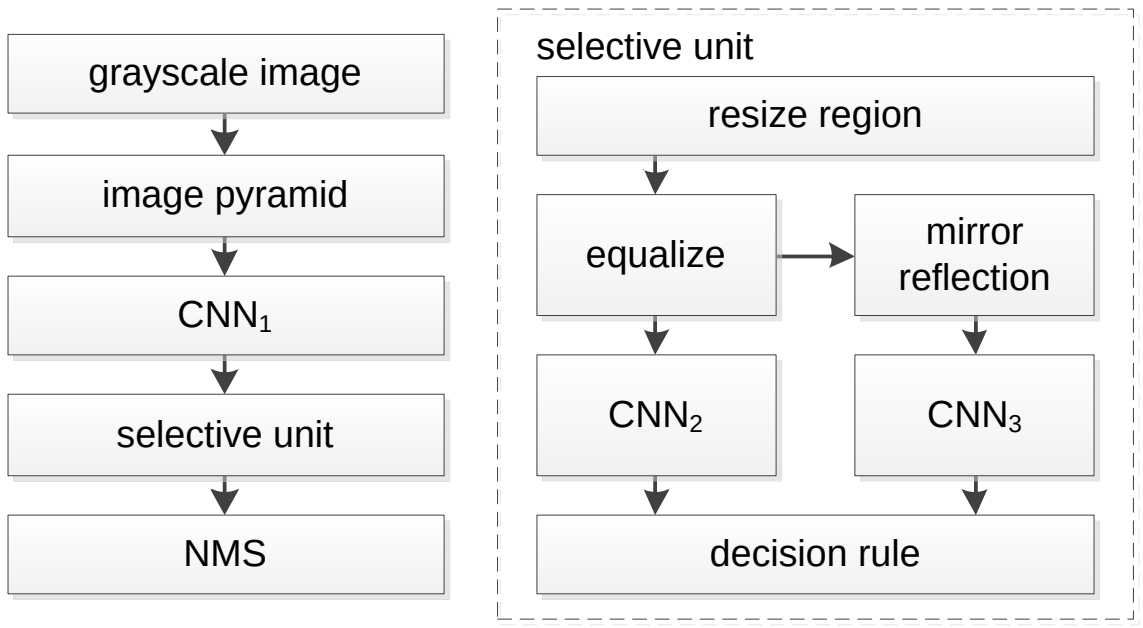

Figure 3: Detector design.

## 3.4. Implementation

The detector has been implemented using three technologies: SIMD expansion of x86 processor family (for each of three instructions sets – SSE, AVX, AVX2 – supported by microarchitectures of Intel Sandy/Ivy Bridge and Haswell/Broadwell processors), Nvidia CUDA, OpenCL. The calculations are carried out using single precision, and a precision-recall characteristic is identical for all implementations.

Thus, the implementations of all stages of the pipeline detector (except for NMS) are presented in several versions of manually vectorized code. We used vector intrinsic functions (which are directly translated into the appropriate assembly instructions by the compiler) and took into account the limited number of logical registers, as well as minimized the number of queries to the memory. At the same time, the SSE code can be ported to the ARM platform, as all used SSE instructions have analogs in the NEON set of instructions. Scalar C++ code and OpenCL allows you to run the detector on the most of devices, but with less execution speed.

Figure 4 shows the comparison of our convolution implementation with its implementation in the Intel IPP, Nvidia NPP and cuDNN, ArrayFire libraries. Time measurements were made for the first $CNN_1$ convolution layer calculation (Fig. 1) on the image with a 4K Ultra HD resolution and were averaged over $10^3$ launches. When implementing the convolution for GPU, we used the method proposed by F. N. Iandola et al. [7]. Due to the fine code optimization for each CNNs architecture, the speed of layers calculation is higher than when using more universal functions from the respective libraries.

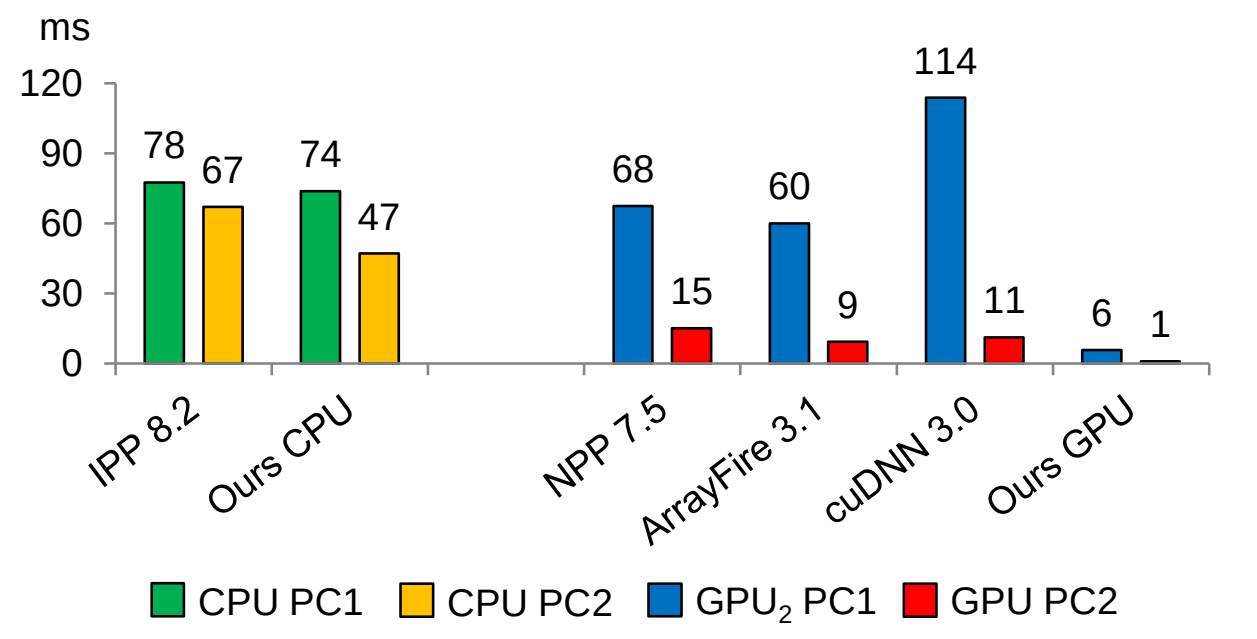

Figure 4: Comparison of the calculation speed of the first convolution layer for $CNN_1$ for an image with a 4K Ultra HD resolution (equipment specifications are listed in section 4).

In order to provide continuous GPU load, scanning of several levels of the image pyramid (3 by default) is executed simultaneously in different streams (concurrent kernels). However, a kernel computation start is synchronized between streams, as they use common pointer to texture memory.

Also, we implemented several solutions to improve the detection speed.

*Asynchronous mode*. Typically, face images occupy only a small area of the image. The main advantage of cascade detectors is the ability to rapidly reject the majority of background regions already in the first stages. However, complex non-facial images may be rejected only in the latter stages. If there is a large number of stages, the speed of detector becomes strongly dependent on the structure of the background and the number of the objects in the scene. At the same time, the non-uniform distribution of the processing load to the image area negatively affects the runtime efficiency of GPU implementations.

To solve this problem, we proposed the asynchronous mode of cascade execution. In this mode, the first stage, run on GPU, sequentially scans each level of the image pyramid. Coordinates of found regions are transmitted to CPU, which operates as a selective unit. The $CNN_1$ moves to scanning of the next level of the pyramid regardless whether the found regions analysis on CPU has been finished or not.

Due to the low-level optimization, the selective unit carries out a single region analysis for 0.1 ms on a single core processor (CPU PC2, for AVX code). Since through the $CNN_1$ passes 0.01% only (in average) of all window positions, by the time scanning of the last level of the image pyramid is complete, the majority of found regions will have already been processed on CPU. A similar situation is possible under the condition of asynchronous execution of the cascade on different CPU cores. Thus, the CNN cascade is able to provide constant frame processing time, dependent only on frame resolution and productivity of the first stage execution.

*Hybrid mode*. The effectiveness of GPU in image processing is significantly higher than CPU. However, if small images (less than 0.01 megapixels) are considered, the calculation time is limited by delays in running of the kernels. To improve search speed of large size faces we made it possible to run the first stage of the cascade on CPU and GPU simultaneously. In the hybrid mode, CPU begins scanning of the image pyramid from the upper level, while GPU processes a high-resolution image on the lower level. Also, it is possible to use an asynchronous mode. A similar technique was used in [21].

*Patchwork mode*. In order to reduce the kernel launch delays to minimum, it's possible to use a patchwork technique [6]. We applied the Floor-ceiling no rotation (FCNR) algorithm [20] for the dense image pyramid packing in semi-infinite strip of a predetermined width. Thus, scanning of all image scales at once is carried out in a single run of the $CNN_1$. However, it is not possible to use the hybrid mode. In this case, the second stage of the cascade is also performed on GPU by scanning all the found regions in a single pass. Usually the asynchronous mode is more effective than the patchwork mode, but the latter improves the low-resolution image processing performance.

# 4. Experiments

In this section, the results of the proposed detector testing on two public face detection benchmarks – FDDB [8] and AFW [37] – are shown. Since the detector was designed for frontal face search only, it is obvious that it cannot surpass the multi-view face detectors with these complex datasets. However, it is comparable with the state-of-the-art frontal face detectors on the FDDB benchmark.

In addition, we tested several face detectors, which source code or demo versions are in open access. In order to compare performance and speed of algorithms for video processing tasks, tests were carried out on the annotated video.

This section also provides execution speed of all detector implementations for different video stream resolutions and asynchronous mode demonstration. The test results show that the CNN cascade provides very high data processing speed on both GPU and CPU, outperforming all previously proposed algorithms.

Specification of the used equipment:

– PC1 (laptop): Intel Core i7-3610QM CPU (2.3 GHz, Turbo Boost disabled), Intel HD Graphics 4000 $GPU_1$ (GT2, 16 core, 1,100 MHz) and Nvidia GeForce GT 640M $GPU_2$ (GK107, 384 core, 709 MHz);

– PC2 (desktop): Intel Core i5-3470 CPU (3.2 GHz, 3.6 GHz with Turbo Boost) and Nvidia GeForce GTX 960 GPU (GM206, 1,024 core, 1,228 MHz).

## 4.1. Face Detection Data Set and Benchmark

The FDDB [8] benchmark consists of 2,845 pictures (no more than 0.25 megapixels), and it has elliptical shape annotations for 5,171 faces. The authors provide a standardized algorithm for the automatic ROC curves construction based on the detector operation results. The algorithm calculates two types of evaluations: the discrete score and continuous score. ROC curve for the discrete scores reflects the dependence of the detected faces fraction on the number of false alarms by varying the threshold of the decision rule. The detection is considered to be positive if the Intrsection-over-Union (IoU) ratio of detection and annotation areas exceeds 0.5. Only one detection can be matched with annotation. Continuous score reflects the quality of the face localization, i.e. average IoU ratio.

The result of the offered detector evaluation is shown in Figure 5. The following search settings were used: the minimum object size (*minSize*) – 15×15 pixels, the scale factor for the image pyramid construction (*scaleFactor*) – 1.05, $T_1 = 0$, $T_2 = 0$, $T_{nn} = 1$. Since the detector localizes rectangular areas, in some cases this leads to errors when they are matched with the elliptical shape annotations. For a correctly evaluation, we manually modified 105 received bounding boxes (Fig. 6) so that their IoU ratio would exceed a predetermined threshold.

Based on the adjusted evaluation, the performance of our detector exceeds the performance of SURF [17], PEP-Adapt [14] and Pico [18] frontal detectors, approaching the multi-view SURF [17] detector performance. Table 1 shows the average number of the sliding window positions, selected by each CNN cascade stage on the images from the FDDB collection. Even when $T_1 = 0$, more than 99.99% of window positions rejected already at the first stage. That is substantially better than the result obtained in [16].

## 4.2. Annotated Faces in the Wild

The Annotated Faces in the Wild (AFW) [37] benchmark consists of 205 large-scale (0.5-5 megapixels) images and contains rectangular annotations for 468 faces.

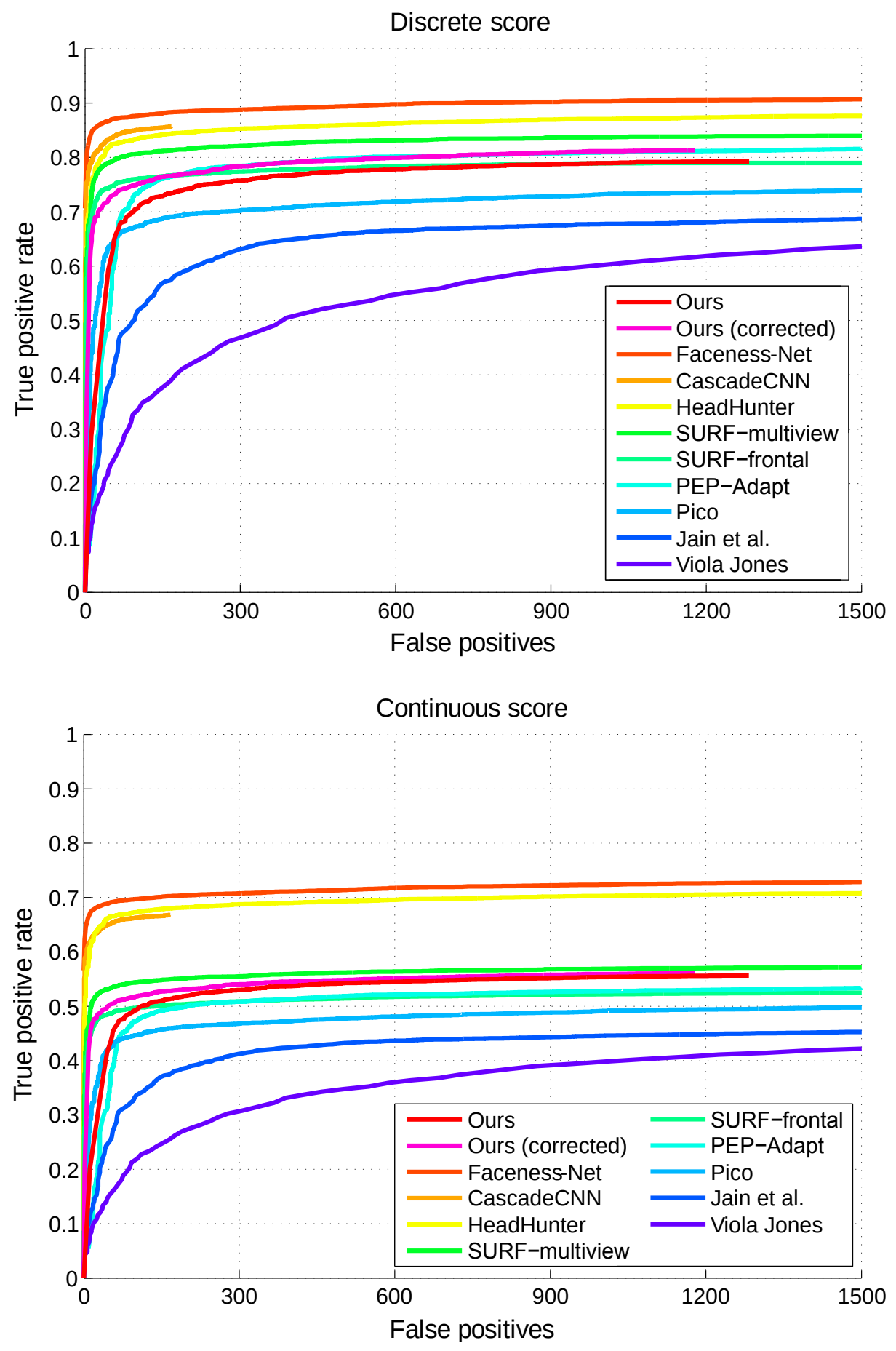

Figure 5: Discrete score ROC curves and Continuous score ROC curves for different methods on FDDB dataset including: our CNN cascade, Faceness-Net [34], CascadeCNN [16], HeadHunter [19], SURF [17], PEP-Adapt [14], Pico [18], Jain et al. [9], Viola Jones (OpenCV).

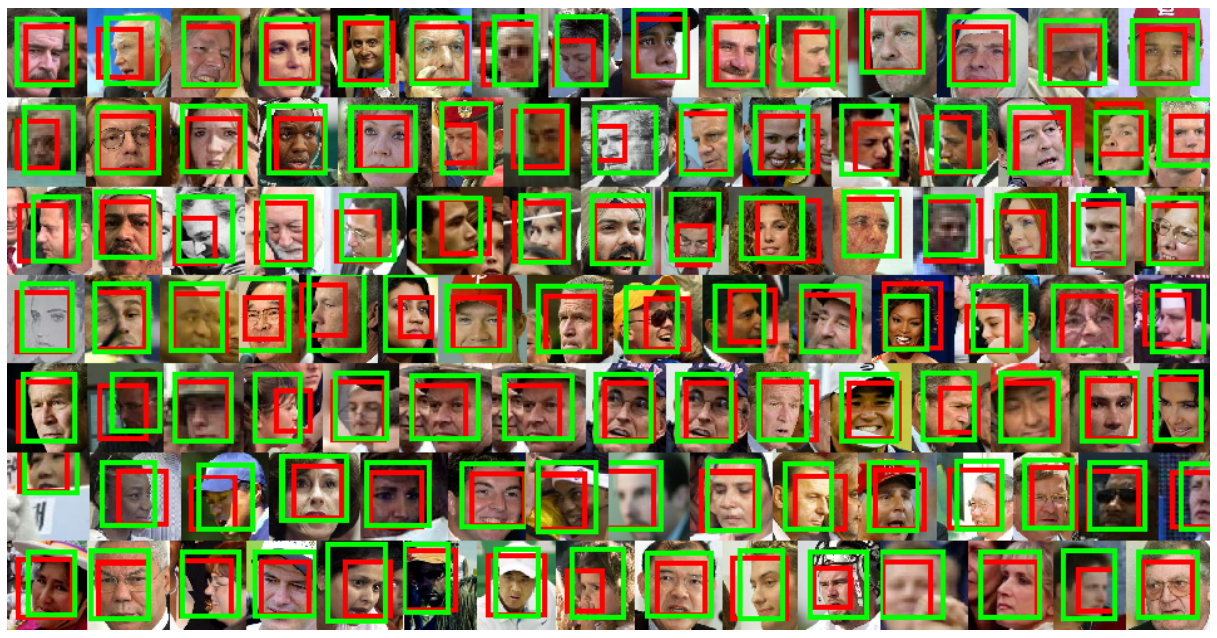

Figure 6: Manually corrected detections received by the CNN cascade on the FDDB benchmark.

This benchmark was developed relatively recently and is mainly used for the multi-view detectors evaluation. Because of it, we additionally tested 7 frontal face detectors, including: two Haar cascades (OpenCV-default, OpenCV-alt) and LBP cascade (OpenCV-lbp) from the OpenCV library; Haar cascade [25] (OpenCV-Pham), and LBP cascade [11] (OpenCV-Köstinger), trained by the OpenCV object detection framework; SURF cascade [17] (SURF-frontal, not the same model as for the FDDB); cascade of decision trees, using the pixels intensity comparison [18] (Pico).

Table 1. Statistics of the CNN cascade operation on the FDDB. It shows the detections number averaged over all images made by each cascade stage, as well as the percentage of rejected windows.

| stages | number windows | rejected windows, % |
|---|---|---|
| sliding window | 2 724 768.2 | |
| stage 1 | 132.7 | 99.995 |
| stage 2 | 57.0 | 57.019 |
| stage 3 | 43.3 | 24.036 |
| NMS | 1.9 | |

For each detector precision and recall scores were calculated for different values of *minNeighbors* = {1, 2, 3} (parameter specifying how many neighbors each candidate rectangle should have to retain it), and the mean value of $F_1$ measure. *MinNeighbors* parameter is used in OpenCV and SURF detectors and is equivalent to $T_{nn}$ (2) for our detector. In Pico implementation, the level of false alarms is regulated by the threshold of the decision rule, which in this test was assumed to be equal to *minNeighbors* + 2.

Comparison of the detectors was carried out with the following search settings: *minSize* = 80×80 pixels, *scaleFactor* = 1.1. SURF and OpenCV-Köstinger detectors localize smaller face area in comparison with other detectors, that is why their *minSize* value was reduced by 25%. Configuration of the detectors:

– OpenCV: version 3.0.0, *useOptimized* = 1;
– SURF: *modelType* = 0, *fast* = 1, *step* = 1;
– Pico: *strideFactor* = 0.1.

Standard IoU ratio with a threshold of 0.5 was used to evaluate the detections. Moreover, 44 bounding boxes were additionally generated for each annotation by scaling it with a factor of 0.9 to 1.2 (Fig. 7). Such multiple check of detections made it possible to take into account differences in the size of areas, localized by detectors, and eliminate error matching.

The test results are presented in Figure 8. The proposed detector ($F_1$ = 0.75) holds the second position in terms of $F_1$ measure value. It is second to detector OpenCV-Köstinger ($F_1$ = 0.78) due to more false alarms. However, on this benchmark, modern multi-view detectors show significantly better results, but they cannot work in real-time mode. For example, the Faceness-Net [34] detector requires 50 ms for image analysis with VGA resolution on Nvidia Titan Black GPU, which is about 200 times more than working time of our CNN cascade.

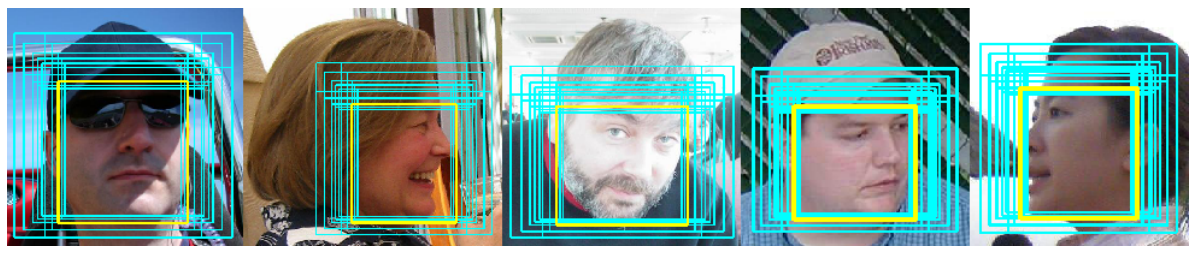

Figure 7: AFW annotations modification.

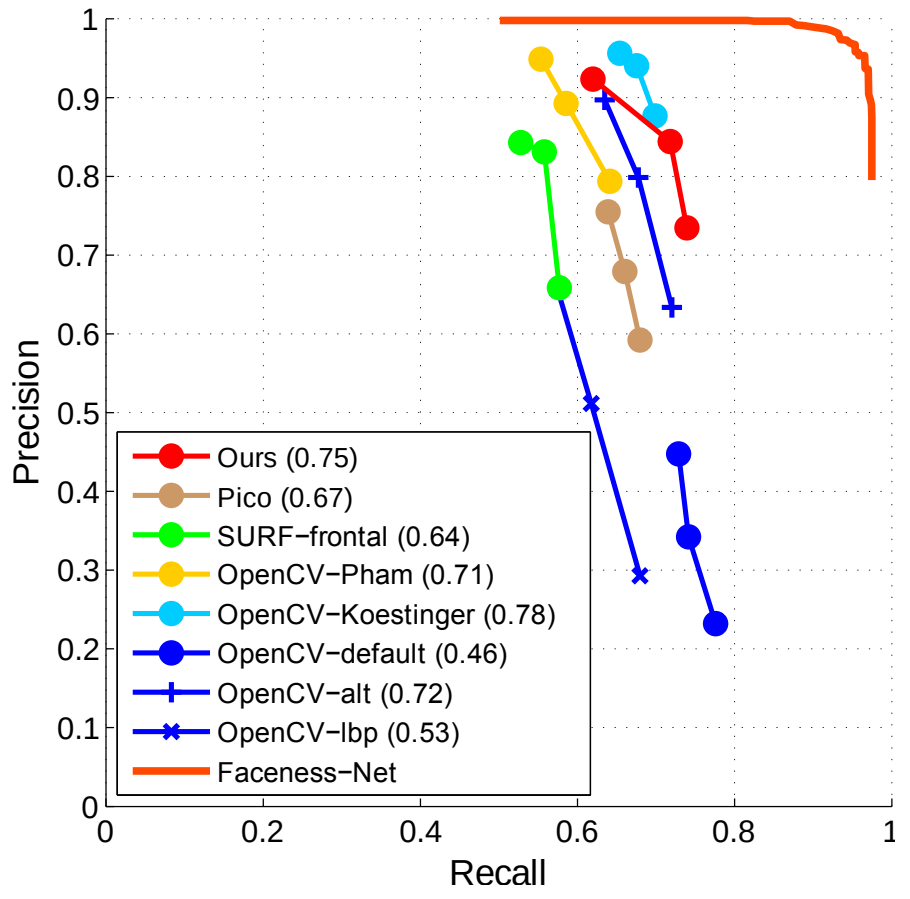

Figure 8: Test results of the frontal face detectors on the AFW benchmark. The numbers in parentheses are mean values of the $F_1$-measure for detectors.

## 4.3. Video data

During the last test, performance of detectors was evaluated on a high-resolution video sequence. We annotated the first 12,000 frames of the 12th episode from the 5th season of «How I Met Your Mother» TV series in HD resolution. This video segment comprises 10 different scenes and the number of faces in a frame varies from 1 to 88 (39,976 faces in total).

For testing, we used the same detector parameters, as for the AFW benchmark, except *minSize* parameter, which was equal to 40×40 pixels. Also, in this test the parameter $T_2$ for our detector was equal to 1.7. All detectors were running on CPU PC1 in a single-threaded mode.

The proposed detector also ranks second in terms of $F_1$ measure value ($F_1$ = 0.61, 10.6 fps), yielding to OpenCV-Köstinger ($F_1$ = 0.65, 1.7 fps), but it is superior to all the detectors in terms of speed (Fig. 10). Thus, the CNN cascade provides the best performance/speed ratio in comparison with boosting cascades.

A greater level of recall, with significantly reduced precision, can be achieved by using a weaker decision rule in the selective unit:

$$\delta = K_{nn}^{CNN2} \geq T_{nn} \vee K_{nn}^{CNN3} \geq T_{nn} \quad (3)$$

A greater level of precision can be achieved by performing an additional validation of detections with Haar cascade OpenCV-alt. Haar cascade was used only after the NMS algorithm, and the detections were pre-scaled to the size of 80×80 pixels. This made it possible to keep high-speed video processing.

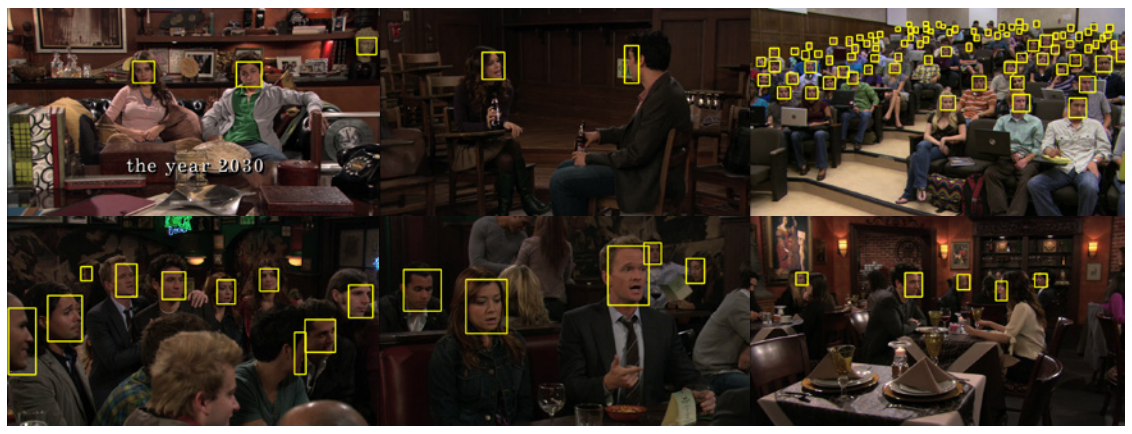

Figure 9: Examples of video frame annotations.

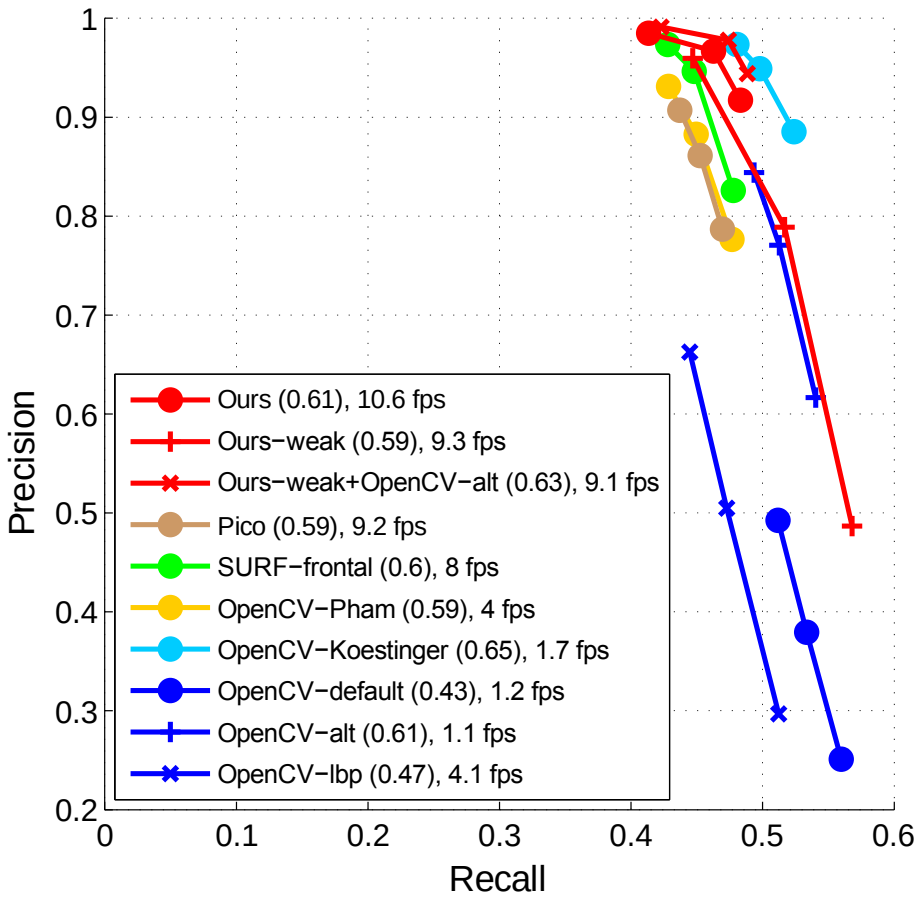

Figure 10: Test results of the frontal face detectors on HD video. The numbers in parentheses are mean values of the $F_1$-measure for detectors.

## 4.4. Runtime efficiency

The runtime efficiency was the key priority at all stages of the proposed detector development. Our CNN cascade turned out to be 2.6 to 10 times faster than Haar and LBP cascades from speed-optimized OpenCV 3.0 library when processing HD video on CPU PC1 (Fig. 10). Among other open source projects (for example, http://libccv.org and http://dlib.net) we do not know a CPU-based algorithm faster than Pico [18] (integer, it does not require the construction of the image pyramid) and SURF cascade [17] (5 stages, SIMD optimization is used). However, despite the higher computational complexity, the speed of the CNN cascade execution exceeds the speed of these detectors due to code vectorization and efficient use of the processor’s cache memory.

For algorithms that have been tested on the FDDB benchmark, the authors reported the following data on the detection speed of frontal faces for images with VGA resolution: NPD[1] [13] – 178 fps (40×40, 1.2, i5-2400 CPU, 4 threads); ACF [35] – 95 fps (i7 CPU, 4 threads); SURF [17] – 91 fps (40×40, 1.2, i5-2400 CPU, 4 threads); Joint Cascade [2] – 35 fps (80×80, 2.93 GHz CPU); Pico [18] – 417 fps (100×100, i7-2600 CPU, 1 threads); Fast DMP [33] – 42 fps (Intel X5650 CPU, 6 threads). The speed of other detectors does not exceed 10 fps. Only methods based on the CNN [3, 16, 34], and HeadHunter [19], support GPU. For typical search settings (*minSize* = 40×40 pixels, *scaleFactor* = 1.2) the CNN cascade guarantees 85 fps for single-threaded and 148 fps for multi-threaded execution on CPU PC2, 171 fps on $GPU_2$ PC1 and 313 fps on GPU PC2 (used hybrid mode).

CNN cascade proposed by H. Li et al. [16] processes a VGA image for 110 ms on CPU (80×80, 1.414, Intel Xeon E5-2620, 1 thread) and for 10 ms on GPU (Nvidia GeForce GTX TITAN Black, 2,880 CUDA core). With the similar search settings, our CNN cascade finishes its operation for 2.5 ms on a single core CPU PC1 and 2 ms on $GPU_2$ PC1.

Many investigations are focused on optimization of Viola-Jones algorithm for GPU with the purpose to increase processing speed of Full HD video stream: D. Oro et al. [22] – 35 fps for Haar cascade (24×24, 1.1, Nvidia GTX470 GPU); S.C. Tek and M. Gokmen [28] – 35 fps for MCT cascade (24×24, 1.15, Nvidia GTX580 GPU); C. Oh et al. [21] – 29 fps for LBP cascade (30×30, 1.2, Nvidia Tegra K1 GPU + Cortex-A15 CPU). Using the same search parameters, the CNN cascade speed reaches up to 75, 98 and 108 fps respectively when it runs in the hybrid mode on PC2. Thus, the proposed detector provides higher runtime efficiency on GPU, in comparison with to Viola-Jones cascade detectors.

Figure 11 shows the dependence of the speed of various detectors on the content of the scene for the first 4,000 frames of the annotated video. When using the asynchronous mode, the CNN cascade provides nearly constant processing time on both CPU and GPU, even when there is a significant increase in the number of faces in the scene (88 faces in frames from 1,771 to 1,834). This property is important for the video analytics systems, as it makes possible to predict more precisely the speed of the system in different use cases.

Figure 12 shows a diagram of an average video frame rate in 4 standard resolutions, which can be reached with different detector implementations optimized for CPU and GPU. Testing was conducted on first 4,000 frames of the annotated video that was scaled to the size of the appropriate resolution. The results indicate that the CNN cascade copes even with the extreme task of real time processing of the video stream with a 4K Ultra HD resolution. For example, speed of the LBP cascade face detector from OpenCV library reaches 3 fps only on $GPU_2$ PC1 with the same search settings.

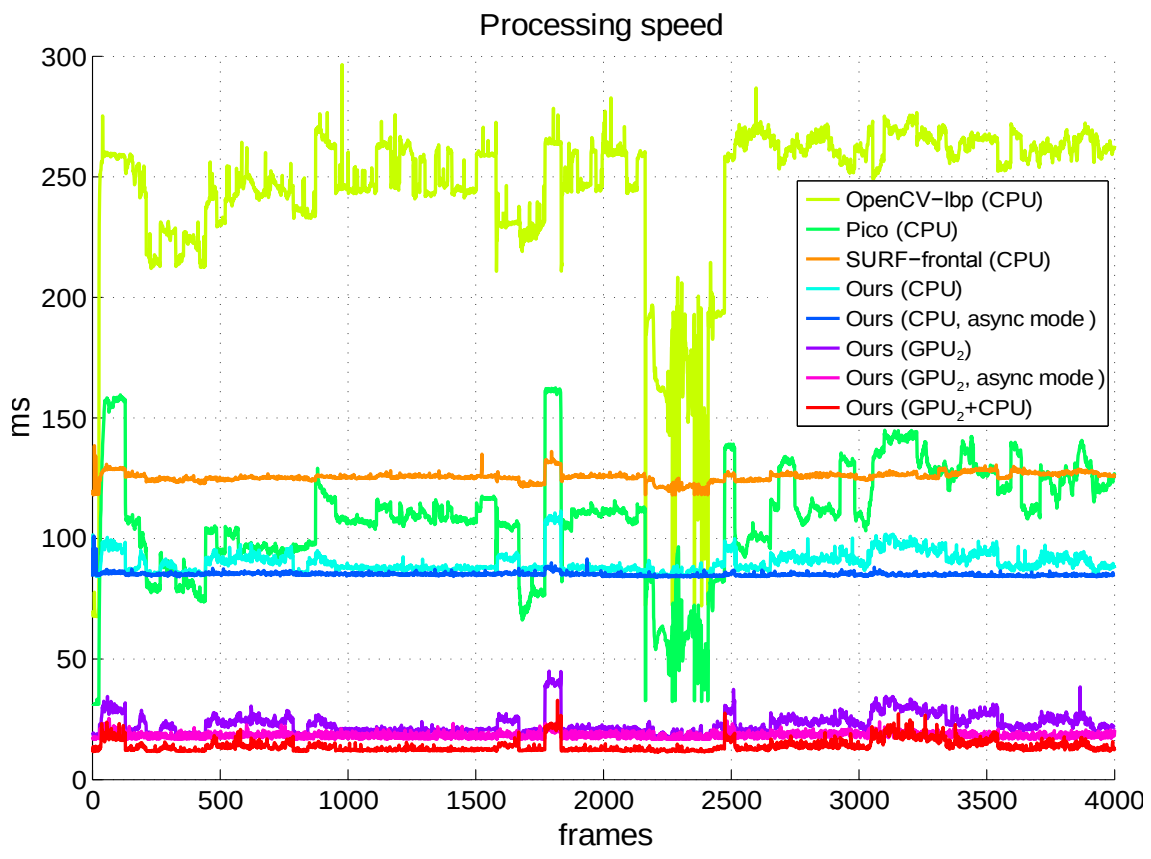

Figure 11: Dependence of detector operation speed on the scene content during HD video processing (search settings the same as in section 4.3, single-threaded execution on PC1).

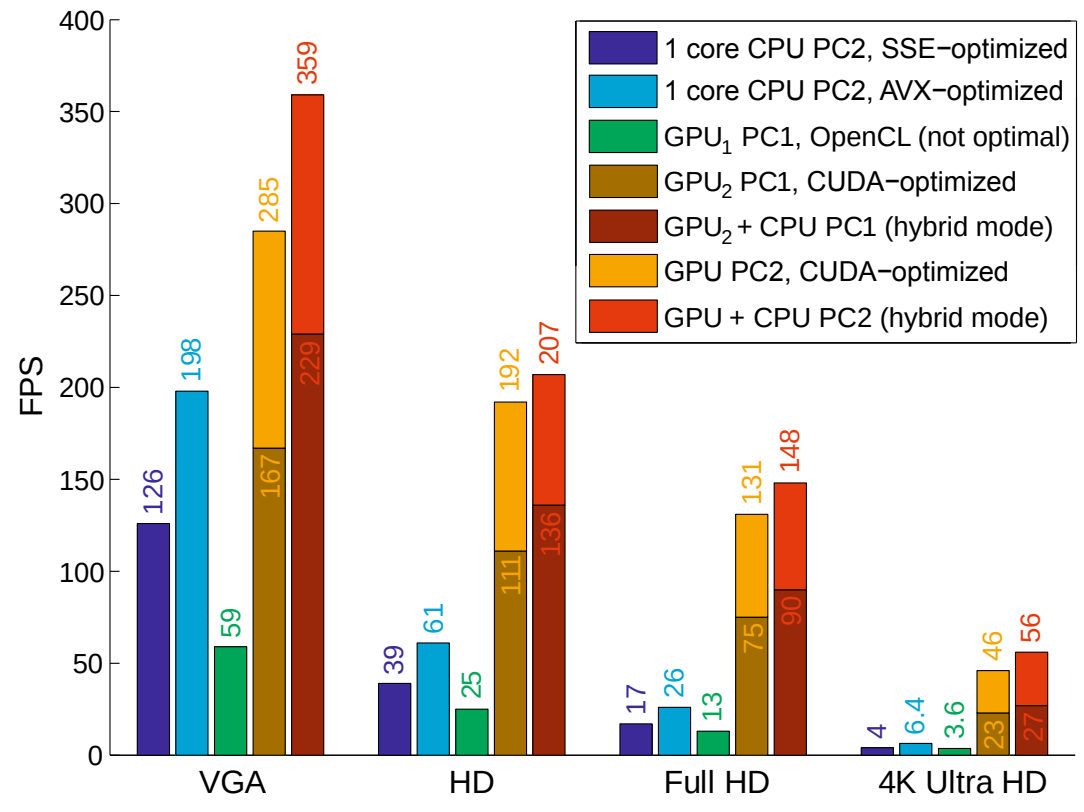

Figure 12: Performance of different frontal face detector implementations based on the compact CNN cascade (*minSize* = 60×60 pixels, *scaleFactor* = 1.2, *minNeighbors* = 2).

## 5. Conclusion

In this paper, we proposed a cascade of compact CNNs for the rapid detection of frontal faces in HD video stream. The first stage of the cascade is capable to reject 99.99% of windows containing background. That, in combination with asynchronous execution mode, substantially reduces the dependence of the detector speed on the image content.

The CNN cascade performance is comparable with the best frontal face detectors on the FDDB benchmark, but it surpasses them in speed on both CPU and GPU. Thus, the proposed algorithm establishes a new level of performance /speed ratio for the frontal face detection problem and makes it possible to get acceptable processing speed even on low-power computing devices.

[1] just 39 fps when processing our annotated video (scaled to the VGA resolution) for multi-threaded execution on CPU PC1.